\documentclass{article}
\usepackage[utf8]{inputenc}
\usepackage{mathtools, nccmath}
\usepackage{geometry}
\usepackage{amsmath}
\usepackage{float}
\usepackage{amssymb}
\usepackage{bbm}
\usepackage{bbold}
\DeclareMathOperator*{\argmax}{arg\,max}

\title{A Comparison of Reward Functions in Q-Learning Applied to a Cart Position Problem}
\author{Amartya Mukherjee$^1$\\\small{$^1$Department of Applied Mathematics, University of Waterloo}}
\date{May 2021}

\begin{document}

\maketitle

\begin{abstract}
    Growing advancements in reinforcement learning has led to advancements in control theory. Reinforcement learning has effectively solved the inverted pendulum problem \cite{RLIntro} and more recently the double inverted pendulum problem \cite{DoubleIP}. In reinforcement learning, our agents learn by interacting with the control system with the goal of maximizing rewards. In this paper, we explore three such reward functions in the cart position problem. This paper concludes that a discontinuous reward function that gives non-zero rewards to agents only if they are within a given distance from the desired position gives the best results.\\
    
\end{abstract}

\section{Introduction}

Reinforcement Learning (RL) is a branch of Machine Learning that deals with agents that learn from interacting with an environment. This is inspired by the trial-and-error method of learning dealt with in psychology \cite{RLIntro}. The goal of any RL problem is to maximize the total reward, and the reward is calculated using a reward function. At any point, the goal of an RL agent is to choose an action that maximizes not only its immediate reward, but also the total reward it expects to get if it follows a certain sequence of actions. The RL algorithm used in this paper is Q-Learning. It is a basic algorithm used in RL, and the math used in it is easy to understand for a reader that does not have much machine learning background. It also comes with the merit that training a Q-Learning algorithm is significantly faster compared to other commonly used RL algorithms.

The cart position problem is a toy problem that will be explored in this paper. The goal is to move a cart from a position $x=0$ to a position $x=r$. This will be done by adjusting the voltage of the cart at every time step. Currently there exists theoretical input functions that solve the cart position problem. It will be interesting to see how RL compares with these functions.

The objective of this paper is to compare three different reward functions based on the performance of the RL agent. The comparison is done in three ways. First, we see which RL agent reaches steady-state motion the earliest. Second, we see which RL agent has the lowest variability in its steady-state motion. Third, we see which RL agent is the closest to $x=r$ in its steady-state motion.

This paper is organized as follows. In section 2, we describe the cart position problem. In section 3, we present the RL algorithm and the reward functions we intend to use. In section 4, the theoretical solution to the cart position problem is explained. In section 5, we present the results of the RL algorithm and compare them. In section 6, we discuss the applicability of the RL algorithm in real life situations. 

\section{Cart Position Problem}

This paper concerns the control of the position of a cart by a rotary motor. Let $x(t)$ be the position of a cart at time $t$ and $V(t)$ the motor voltage at time $t$. $\eta_g$ is the gearbox efficiency, $K_g$ the gearbox gear ratio, $\eta_m$ the motor efficiency, $K_t$ the motor torque constant, $r_{mp}$ the motor pinion radius, $K_m$ the back-EMF constant, $R_m$ the motor armature resistance, $\nu$ the coefficient of viscous friction, and $M$ the mass of the cart. These are all constants. The equation for this control system is a second-order differential equation shown in equation \ref{eq:Main} below:

\begin{equation}
    M\ddot{x}(t)=\frac{\eta_gK_g\eta_mK_t(r_{mp}V(t)-K_gK_m\dot{x}(t))}{R_mr_{mp}^2}-\nu\dot{x}(t)
    \label{eq:Main}
\end{equation}

Throughout the course of this paper, we can define constants $\alpha$ and $\beta$ as:
$$\alpha=-\frac{\eta_gK_g^2\eta_mK_tK_m}{MR_mr_{mp}^2}-\frac{\nu}{M}$$
$$\beta=\frac{\eta_gK_g\eta_mK_tr_{mp}}{MR_mr_{mp}^2}$$

This simplifies our governing equation to equation \ref{eq:MainSimplified} below:

\begin{equation}
    \ddot{x}(t)=\alpha\dot{x}(t)+\beta V(t)
    \label{eq:MainSimplified}
\end{equation}

Where $\alpha<0$ and $\beta>0$. In the absence of any control input $V(t)$, the solution to the differential equation is:

$$x(t)=C_1e^{\alpha t}+C_2$$

In this model, $\alpha<0$, so $\lim_{t\to\infty}x(t)=C_2$. The steady-state position is $C_2$. The goal of the input function $V(t)$ is to change the steady-state position to a position of our choice.

We can express our governing equation as a system of first order differential equations. Let $s(t)=\dot{x}(t)$ represent the velocity of the cart. The system of first order equations is expressed below:

$$\dot{x}(t)=s(t)$$
$$\dot{s}(t)=\alpha x(t)+\beta V(t)$$

Let $X(t)=\begin{bmatrix}x(t)\\s(t)\end{bmatrix}$. This system of first order equations can be expressed in matrix form, as shown below:

$$\dot{X}(t)=\begin{bmatrix}0&1\\0&\alpha\end{bmatrix}X(t)+\begin{bmatrix}0\\\beta\end{bmatrix}V(t)$$

We also know that the objective of our control problem is to drive the cart to a particular position. For this reason, we want our output $y(t)$ to be the position $x(t)$. To express this in a matrix form involving $X(t)$, this can be written as:

$$y(t)=\begin{bmatrix}1&0\end{bmatrix}X(t)$$

The state-space form of our equation is:
\begin{subequations}
\begin{equation}
    \dot{X}(t)=\begin{bmatrix}0&1\\0&\alpha\end{bmatrix}X(t)+\begin{bmatrix}0\\\beta\end{bmatrix}V(t)
    \label{eq:StateSpace1X}
\end{equation}
\begin{equation}
    y(t)=\begin{bmatrix}1&0\end{bmatrix}X(t)+[0]V(t)
    \label{eq:StateSpace1Y}
\end{equation}
\label{eq:StateSpaceEqns}
\end{subequations}

Where $X(t)$ is the state, $V(t)$ is the input, $y(t)$ is the output. The objective of the cart position problem is to find an input function $V(t)$ so that the steady-state value of $x$ is $r$. And this will be done using RL.

\section{Use of RL in the Cart Position Problem}

The initial conditions used in the control system are: $x(0)=\dot{x}(0)=0$. In this paper, we will let $\alpha=-1,\beta=10,r=10$. The solution to Equation \ref{eq:StateSpaceEqns} is computed numerically using Euler's method with a step-size of $\delta t=0.2$s.

At each time step, the RL agent takes the output of the control system, which is the position $x(t)$ of the cart, and returns the voltage $V(t)$ that should be used as the input to the control system at that time step. The voltage $V(t)$ is an integer between $-5$ and $5$. In order to ensure that the cart does not go too far from $x=r$, the bounds of $x$ has been set to $[-10,20]$.

The training process involves running $100$ samples. Each sample has $30$ rounds. When a round starts, the cart is at position $x=0$. The round ends if the cart goes out of bounds (i.e. $x\not\in [-10,20]$) or $50$ time steps have passed since the start of the round. This means the training process of the RL agent involves observing at most $150000$ time steps.

The objective of the RL agent is to find a sequence of voltage inputs such that the total reward is maximized.

\subsection{Q-Learning}

The RL algorithm used in this paper is Q-Learning. Q-Learning was first coined by Watkins (1989) in his PhD thesis \cite{Watkins1989}. It finds an optimal policy $Q(s,a)$ where $s$ is the state and $a$ is the action. This optimal policy maximizes the total reward in the system. $Q(s,a)$ can be thought of a table of values for all states and actions.

At first, $Q(s,a)$ is initialized to $0$ for all $s,a$. At every time step $t$, the action $a_t$ is determined from the state $s_t$ through an $\epsilon$-greedy process. In order to find the best action $a_t$ from $Q(s,a)$, we must first use a random process to figure out the result of the action. When the training of the agent starts, we let $\epsilon=1$. At every time step, we take a random sample from the $Uniform(0,1)$ distribution. If the random sample is greater than $\epsilon$, then $a_t=\argmax_a(Q(s_t,a))$. Otherwise, $a_t$ is a randomly selected action.

Let $Q^{[n]}(s,a)$ be the policy after $n$ updates. When $s_{t+1}$ is calculated using $s_t$ and $a_t$, then $Q^{[n]}(s,t)$ is updated using the following equation:

\begin{equation}
    Q^{[n+1]}(s_t,a_t)=Q^{[n]}(s_t,a_t)+\zeta(r_t+\gamma\max_aQ(s_{t+1},a)-Q(s_t,a_t))
\end{equation}

Where $r_t$ is the reward, $\zeta$ is the learning rate. $\max_aQ(s_{t+1},a)$ is a measure of the guess of the future reward, and $\gamma$ is a measure of the weight we give to future rewards. $\zeta$ and $\gamma$ are hyperparameters to this agent. The hyperparameters used in this paper are: $\zeta=0.05,\gamma=0.9$

In $Q(s,a)$, the state $s$ refers to the position $x$ of the cart, and the action $a$ refers to the voltage $V$ of the cart.

The Q-Learning code used in this paper is taken from Lin's (2018) GitHub repository. This code was originally used to solve the Blackjack problem using Q-Learning. The equation for updating $\epsilon$ after every time step $t$ according to Lin's code is given by the equation below \cite{Lin2018}:

\begin{equation}
    \epsilon_{t+1} = \begin{cases}
    \epsilon_t - \epsilon_0/(3n_{train}) & n_{train}-t > 0.7n_{train} \text{ or } 0.3n_{train} > n_{train}-t > 0\\
    \epsilon_t - 2\epsilon_0/n_{train} & 0.7n_{train} > n_{train}-t > 0.3n_{train}\\
    0 & \text{Otherwise}
    \end{cases}
\end{equation}

Where $n_{train}=30000$ is the number of time steps used to train the agent.

\subsection{Reward Functions}

As shown previously, reward functions are used to update the policy function at each time step with the intention that the RL agent maximizes the total reward. This paper will explore three reward functions. The first reward function takes the negative of the square of the distance between $x$ and $r$ with the intention that the RL agent will be rewarded better if it is closer to $r$.

\begin{equation}
    R(x)=-(x-r)^2
    \label{eq:R1}
\end{equation}

The second reward function is piece-wise and is always positive. This function is linear instead of quadratic.

\begin{equation}
    R(x)=\begin{cases}
    \frac{r}{2}-|x-r|&\frac{r}{2}<x<\frac{3r}{2}\\
    0&\text{Otherwise}
    \end{cases}
    \label{eq:R2}
\end{equation}

The third reward function is discontinuous. It only rewards a $1$ if the cart is a distance of less than $1$ unit away from $r$ and it rewards a $5$ if the cart is a distance of less than $0.1$ units away from $r$.

\begin{equation}
    R(x)=\begin{cases}
    1&x\in[r-1,r-0.1]\cup[r+0.1,r+1]\\
    5&x\in(r-0.1,r+0.1)\\
    0&\text{Otherwise}\\
    \end{cases}
    \label{eq:R3}
\end{equation}

To reproduce our results, we provided the code used for this problem \cite{Amartya2021}.

\section{Theoretical solution}

Consider the following voltage function:

\begin{equation}
    V(t)=K_p(r-x(t))
\end{equation}

Where $K_p>0$. Substituting the voltage function into equation \ref{eq:MainSimplified} gives us the following:

\begin{equation}
    \ddot{x}(t)=\alpha\dot{x}(t)+\beta K_p(r-x(t))
    \label{eq:Theoretical2}
\end{equation}

This equation gives us a solution $x(t)$ whose steady-state position is $x=r$ \cite{Farsi2021}. The proof is given in Appendix A. Numerical simulations of Equation \ref{eq:Theoretical2} have been done using Euler's method with $K_p=0.2$ and $K_p=0.1$. The plot of the two trajectories are given below:

\begin{figure}[H]
    \centering
    \includegraphics[width=10cm]{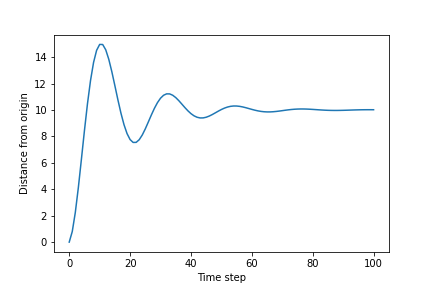}
    \caption{Trajectory of the cart using the theoretical solution with $K_p=0.2$}
    \label{fig:T2}
\end{figure}

\begin{figure}[H]
    \centering
    \includegraphics[width=10cm]{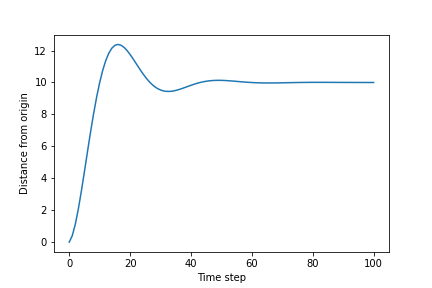}
    \caption{Trajectory of the cart using the theoretical solution with $K_p=0.1$}
    \label{fig:T1}
\end{figure}

With $K_p=0.2$, the position of the cart is within $[r-1,r+1]$ after $35$ time steps (or $7$ seconds).
With $K_p=0.1$, the position of the cart is within $[r-1,r+1]$ after $23$ time steps (or $4.6$ seconds).

This theoretical solution will be compared to the solution derived from Q-Learning.

\section{Results of Q-Learning}

In this section, we train three RL agents using each of the reward functions provided in section 3.2. We then compare the results of each of our RL agents.

\subsection{Reward Function 1}

The first reward function takes the negative of the square of the distance between $x$ and $r$ with the intention that the RL agent will be rewarded better if it is closer to $r$, as shown in Equation \ref{eq:R1}. Figure \ref{fig:RP1} below shows how the average reward varies with each sample. The average reward is calculated by taking the mean of the total reward of each of the trajectories in the sample.

\begin{figure}[H]
    \centering
    \includegraphics[width=10cm]{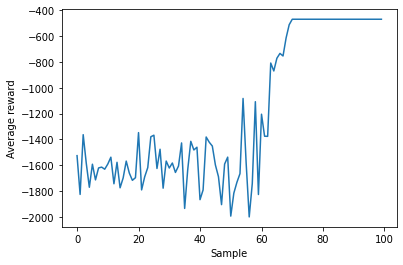}
    \caption{Average reward plot for reward function 1}
    \label{fig:RP1}
\end{figure}

\begin{figure}[H]
    \centering
    \includegraphics[width=10cm]{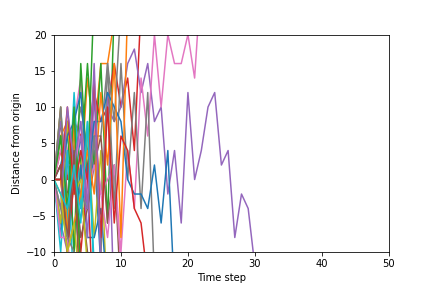}
    \caption{Trajectory of each cart at sample 1 for reward function 1}
    \label{fig:R1S1}
\end{figure}

It is clear that, after training for $70$ samples, the training is complete and the average reward is approximately $-468$. Figure \ref{fig:R1S1} shows the trajectory of each cart in sample 1. These carts follow a random motion. All trajectories of carts in later samples will be compared to this plot.

In this graph, most of the carts go out of bounds within the first 10 time steps. During the first sample, the value of $\epsilon$ used in the $\epsilon$-greedy algorithm is almost $1$, which is why the voltage applied at each time step is picked randomly. Thus, the trajectory that each cart takes is random.

\begin{figure}[H]
    \centering
    \includegraphics[width=10cm]{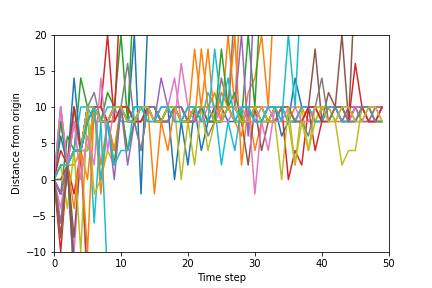}
    \caption{Trajectory of each cart at sample 65 for reward function 1}
    \label{fig:R1S65}
\end{figure}

At samples $60$ to $70$, the average reward plot in Figure \ref{fig:RP1} shows an increasing trend. Figure \ref{fig:R1S65} shows the trajectory of each cart at sample $65$.

In this plot, clearly the RL agent has learned that the cart needs to move towards $x=r$ during the start of the round. The trajectories are less likely to move out of bounds. Several trajectories appear to move back and forth $x=r$. It shows a significant improvement compared to Figure \ref{fig:R1S1}.

\begin{figure}[H]
    \centering
    \includegraphics[width=10cm]{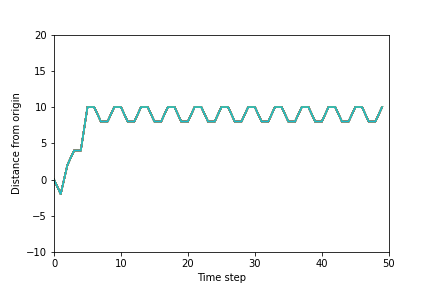}
    \caption{Trajectory of each cart at sample 100 for reward function 1}
    \label{fig:R1S100}
\end{figure}

Figure \ref{fig:R1S100} shows the trajectory of each cart in sample $100$.

This graph shows that the cart reaches a steady-state motion after $5$ time steps (or $1.0$ seconds). The position alters between $x=10.0$ and $x=8.0$ and the voltage alters between $V=-1$ and $V=1$. The total reward in this trajectory is $-468$, which is significantly high compared to the average rewards in the first few samples shown in Figure \ref{fig:RP1}.

The steady-state position is within $\pm 2$ of $r$. If $V=0$ at $x=10$, then the cart is expected to drift past $x=r$, which is why $V=-1$ is the best action to take here.

\subsection{Reward Function 2}

The second function gives a positive reward only if the position of the cart is in the range $[\frac{r}{2},\frac{3r}{2}]=[5,15]$, as shown in Equation \ref{eq:R2}. Contrary to reward function 1 and 3, the step-size used here is $0.1$s because it leads to better results, and the number of time steps in a round is still $50$. Figure \ref{fig:RP2} below shows how the average reward varies with each sample.

\begin{figure}[H]
    \centering
    \includegraphics[width=10cm]{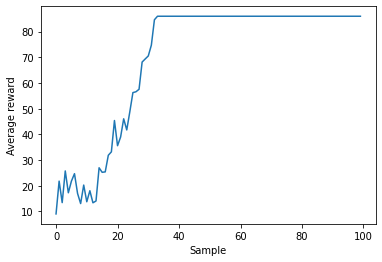}
    \caption{Average reward plot for reward function 2}
    \label{fig:RP2}
\end{figure}

\begin{figure}[H]
    \centering
    \includegraphics[width=10cm]{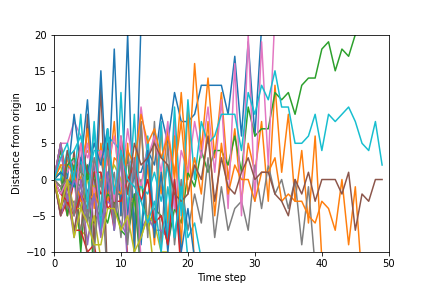}
    \caption{Trajectory of each cart at sample 1 for reward function 2}
    \label{fig:R2S1}
\end{figure}

It is clear that, after training for $40$ samples, each trajectory of the cart has a total reward of $86$. Figure \ref{fig:R2S1} shows the trajectory of each cart in sample 1. These carts follow a random motion. All trajectories of carts in later samples will be compared to this plot. The distance from origin is the value $x(t)$ at time $t$ and the time step is $t/\delta t$.

In this graph, most of the carts go out of bounds within the first 40 time steps. This makes sense since the time step size used here is half the step size used for the first reward function. For this reason, the trajectories shown in this graph are different from the trajectories shown in Figure \ref{fig:R1S1}. The trajectory that each cart takes is random because the voltage applied at each time step is picked randomly.

\begin{figure}[H]
    \centering
    \includegraphics[width=10cm]{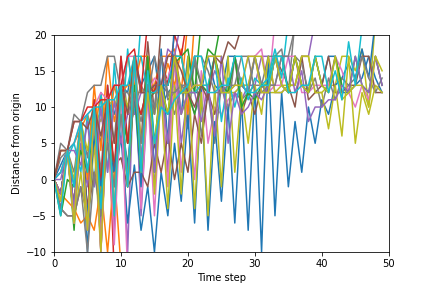}
    \caption{Trajectory of each cart at sample 25 for reward function 2}
    \label{fig:R2S25}
\end{figure}

At samples $20$ to $30$, the average reward in Figure \ref{fig:RP2} shows an increasing trend. Figure \ref{fig:R2S25} shows the trajectory of each cart in sample 25.

In this plot, it is clear the cart has learned to move towards $x=r$ during the start of the round. Several trajectories appear to stay near $x=r$ and the cart is less likely to go out of bounds. It shows a significant improvement compared to Figure \ref{fig:R2S1}.

\begin{figure}[H]
    \centering
    \includegraphics[width=10cm]{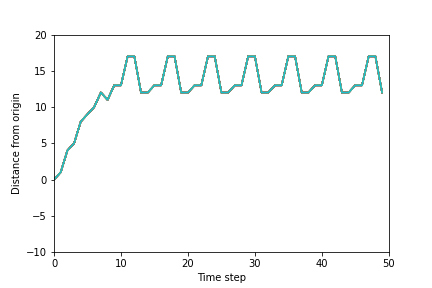}
    \caption{Trajectory of each cart at sample 100 for reward function 2}
    \label{fig:R2S100}
\end{figure}

Figure \ref{fig:R2S100} shows the trajectory of each cart in sample 100.

This graph shows that the cart reaches a steady-state motion after $10$ time steps (or $1.0$ seconds). The validation plot is the same as this plot. The position alters between $x=12.0$, $x=13.0$ and $x=17.0$ and the voltage alters between $V=1$, $V=4$ and $V=-5$. The total reward in this trajectory is $86.0$, which is significantly high compared to the average rewards in the first few samples shown in Figure \ref{fig:RP2}.

However, the variation in the position during the steady-state motion is too big. The steady-state position is within $\pm 7$ of $r$. And $r$ is not in the range of the positions in the steady-state motion. This is because the agent has learned that a positive reward comes if the cart is in the range $[5,15]$. Thus, there is less of an incentive to stay closer to $x=r$ since the current trajectory already significantly maximizes the total reward. This shows that the second reward function is not as reliable as the first reward function.

\subsection{Reward Function 3}

\begin{figure}[H]
    \centering
    \includegraphics[width=10cm]{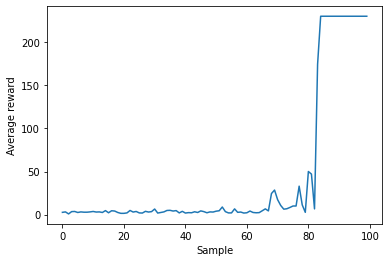}
    \caption{Average reward plot for reward function 3}
    \label{fig:RP3}
\end{figure}

As a measure of ensuring that the variability of the steady-state position is minimized, the third reward function gives a positive reward only if the position of the cart is in the range $[r-1,r+1]=[9,11]$, as shown in Equation \ref{eq:R3}. Figure \ref{fig:RP3} shows how the average reward varies with each sample.

\begin{figure}[H]
    \centering
    \includegraphics[width=10cm]{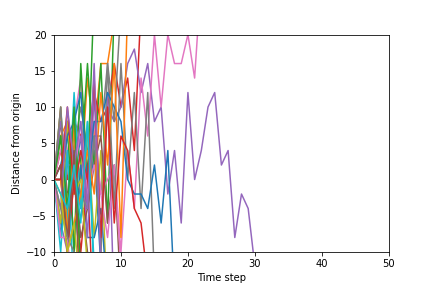}
    \caption{Trajectory of each cart at sample 1 for reward function 3}
    \label{fig:R3S1}
\end{figure}

It is clear that, after training for $85$ samples, each trajectory of the cart has a total reward of $230$. Figure \ref{fig:R3S1} shows the trajectory of each cart in sample 1. These carts follow a random motion. All trajectories of carts in later samples will be compared to this plot.

In this graph, most of the carts go out of bounds within the first 10 time steps. This graph shows similar trends to Figure \ref{fig:R1S1}. The direction each cart goes at the start of the round is random as the voltage applied at each time step is picked randomly.

\begin{figure}[H]
    \centering
    \includegraphics[width=10cm]{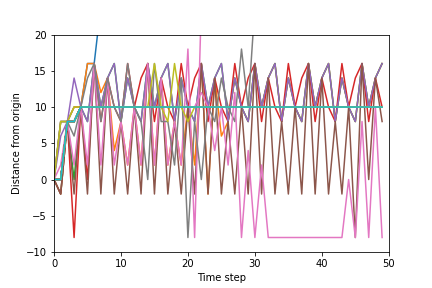}
    \caption{Trajectory of each cart at sample 84 for reward function 3}
    \label{fig:R3S84}
\end{figure}

At samples $80$ to $85$, the average reward in Figure \ref{fig:RP3} shows an increasing trend. Figure \ref{fig:R3S84} shows the trajectory of each cart in sample $84$.

In this plot, it is clear that the agent has learned to move the cart towards $x=r$ during the start of the round. Trajectories are less likely to go out of bounds. In some trajectories, the cart moves back and forth the interval $[r-1,r+1]$ multiple times in order to increase its total reward.

\begin{figure}[H]
    \centering
    \includegraphics[width=10cm]{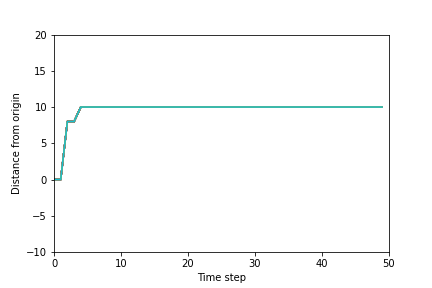}
    \caption{Trajectory of each cart at sample 100 for reward function 3}
    \label{fig:R3S100}
\end{figure}

Figure \ref{fig:R3S100} shows the trajectory of each cart in sample 100.

This graph shows that the cart reaches exactly $x=r$ after $4$ time steps ($0.8$ seconds) and stays there. The validation plot is the same as this plot. The steady-state motion has no variability compared to the first and second reward function. On the other hand, a drawback of using this reward function is that training this agent takes more samples compared to training RL agents that use the first or second reward function, as shown in Figure \ref{fig:RP3}.

We first compare which RL agent reaches steady-state motion the earliest. For the first and second reward function, the Q-Learning agents take equally as long to reach steady-state motion ($1.0$s). The Q-Learning agent using the third reward function reaches steady-state motion the quickest ($0.8$s).

We then compare which RL agent has the lowest variability in its steady-state motion. In the first reward function, the steady-state motion alternates between positions $8.0$ and $10.0$, thus having a width of $2.0$. In the second reward function, the steady-state motion alternates between positions $12.0,13.0$ and $17.0$, thus having a width of $5.0$. In the third reward function, the steady-state motion just takes values of $10.0$, thus having a width of $0.0$. This shows that, while the first reward function leads to a smaller width compared to the second reward function, the third reward function has the smallest width.

Lastly, we compare which RL agent is closest to $x=r$ in its steady-state motion. Clearly, the third reward function performs the best since it's steady state position is exactly $r$. And the second reward function performs the worst since $r$ is not in its set of steady-state positions.

Thus, the third reward function gives the best results.

\section{Discussion}

While the previous section discussed the steady-state behaviour of the Q-Learning agent for each of the reward functions, this section will discuss the practicality of Q-Learning agents on real carts based on the results of the third reward function.

We first compare the trajectory of the Q-Learning agent shown in Figure \ref{fig:R3S100} with the trajectories of the theoretical solutions shown in Figure \ref{fig:T2} and \ref{fig:T1}. It is clear that the Q-Learning agent shows better performance compared to the theoretical solutions as the Q-Learning agent reached steady-state motion within $0.8$ seconds and the theoretical solutions reached a position within $\pm 1$ of $x=r$ within $7$ seconds and $4.6$ seconds.

\subsection{Applicability of Q-Learning on real carts}

In this paper, the cart position problem is thought of as a toy problem where Q-Learning may be useful. Suppose a real cart follows the model shown in Equation \ref{eq:Main} and its goal is to get from a position $x=0$ to $x=r$. This means the position is updated continuously as opposed to time steps shown in Euler's method. In Q-Learning, the policy function $Q(s,a)$ picks actions based on trial-and-error. The set of states $s$ in $Q(s,a)$ are finite and discrete. Thus, Q-Learning is impractical on a real cart where the position can be any real number. In this situation, the theoretical solution is more effective as it adjusts the voltage based on a continuous set of positions.

Having a discrete set of actions can also be undesirable in this problem. In the situation with reward function 1 (Figure \ref{fig:R1S100}), it is clear that $V$ has to alter between $-1$ and $1$ in order to maintain a steady-state motion. If $V=0$ at $x=10$, then the cart will drift away from $x=10$ since its velocity is non-zero. Thus, further areas of exploration involve adding values of $V$ with a smaller magnitude (i.e. $0.1,0.2$) into the range of actions the Q-learning agent could take.

Further areas of exploration also involve using Deep Q learning. Artificial neural networks (ANNs) are trained by modifying weights rather than entries in a $Q(s,a)$ table, thus, ANNs may be more reliable in problems involving continuous states or continuous actions.

Consider the trajectory of the Q-Learning agent shown in Figure \ref{fig:R3S100}. Between time $t=0$ and $t=\delta t=0.2$, the velocity of the cart increases from $\dot{x}=0ms^{-1}$ to $\dot{x}=40ms^{-1}$. This means the acceleration is approximately $\ddot{x}=200ms^{-2}$. In a real cart, this could be dangerously high, given that cars that have an acceleration of approximately $10ms^{-2}$ to $20ms^{-2}$ are considered one of the fastest accelerating cars \cite{Autocar2020}. This shows that the Q-Learning agent is impractical on real carts.

\begin{figure}[H]
    \centering
    \includegraphics[width=10cm]{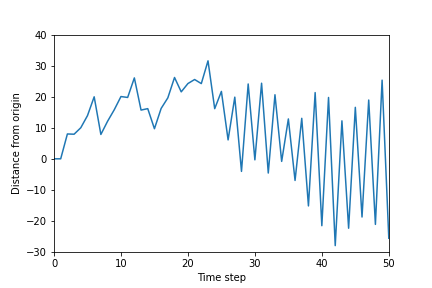}
    \caption{Trajectory of a cart with $\alpha=-1.01$}
    \label{fig:R3T}
\end{figure}

Lastly, in real world carts, there is always an error associated with measuring the mass of the cart, the coefficient of friction, or any of the constants described in Equation \ref{eq:Main}. This means there can be an error associated with measuring $\alpha$ or $\beta$. Suppose we trained a Q-Learning agent with $\alpha=-1$ and we intend to test it on a cart with $\alpha=-1.01$. Figure \ref{fig:R3T} shows what the trajectory will look like.

If we use a cart with $\alpha=-1$, we get the trajectory shown in Figure \ref{fig:R3S100}. If we use a cart with $\alpha=-1.01$, we get the trajectory shown in Figure \ref{fig:R3T}. This shows that a small error in the measurement of $\alpha$ can lead to a significant change in the trajectory of the cart. This occurs because the set of states $s$ in $Q(s,a)$ are finite and discrete. This is why a small change in $\alpha$ means that the cart will reach positions that are either not in the set of states in $Q(s,a)$, or have a different action associated with it. This again shows that Q-Learning is not reliable for real carts.

\section{Conclusion}

This paper compared three different reward functions based on the performance of the RL agent if we intend to use Q-learning to solve the cart position problem. In conclusion, a discontinuous reward function that rewards the RL agent only if the position of the cart lies in $[r-1,r+1]$ gives the best results. Through this analysis, we also created a RL agent that outperforms theoretical solutions.

\section*{Acknowledgements}

The author gratefully acknowledges Jun Liu and Milad Farsi for their continued support, feedback and suggestions on this paper.

\newpage

\section*{Appendix A: Proof of the theoretical solution}

Consider the following input function:

$$V(t)=K_p(r-x(t))$$

Substituting the voltage function into equation \ref{eq:MainSimplified}, gives us the following:

$$\ddot{x}(t)=\alpha\dot{x}(t)+\beta K_p(r-x(t))$$

Re-arranging this gives us a non-homogenous second order differential equation:

\begin{equation}
    \ddot{x}(t)-\alpha\dot{x}(t)+\beta K_px(t)=\beta K_pr
    \label{eq:Q4ODE}
\end{equation}

The goal of this subsection is to find the steady-state position $\lim_{t\rightarrow\infty}x(t)$. This involves finding a general solution to equation \ref{eq:Q4ODE}. This will be done by finding the general solution homogenous version of the equation, $x_h(t)$ and by finding a particular solution to the non-homogenous version of the equation, $x_p(t)$.

\subsection*{A.1: Solve the homogenous equation}

$$\ddot{x}_h(t)-\alpha\dot{x}_h(t)+\beta K_px_h(t)=0$$

Writing this in matrix form with $X_h(t)=\begin{bmatrix}x_h(t)\\\dot{x}_h(t)\end{bmatrix}$ gives the following equation:

\vspace{0.5cm}

$$\dot{X_h}(t)=\begin{bmatrix}0&1\\-\beta K_p&\alpha\end{bmatrix}X_h(t)=:A_hX_h(t)$$

The solution to this equation is:

$$X_h(t)=\exp(A_ht)X_h(0)$$

The eigenvalues of $A_h$ are:

$$\frac{\alpha}{2}\pm\sqrt{\frac{\alpha^2}{4}-\beta K_p}$$

Since $\beta>0$, this means:

$$Re(\sqrt{\frac{\alpha^2}{4}-\beta K_p})<|\frac{\alpha}{2}|$$

And since $\alpha<0$, this means:

$$Re(\sqrt{\frac{\alpha^2}{4}-\beta K_p})<-\frac{\alpha}{2}$$

$$Re(\pm\sqrt{\frac{\alpha^2}{4}-\beta K_p})<-\frac{\alpha}{2}$$

$$Re(\frac{\alpha}{2}\pm\sqrt{\frac{\alpha^2}{4}-\beta K_p})<0$$

This shows that $A_h$ is Hurwitz, and by extension:

$$\lim_{t\rightarrow\infty}X_h(t)=\begin{bmatrix}0\\0\end{bmatrix}$$

Thus, the steady-state value of the homogenous equation is: $x_h=0$.

\subsection*{A.2: Find a particular solution to the non-homogenous equation}
Consider:
$$x_p(t)=r$$

Substituting it into equation \ref{eq:MainSimplified} gives:
$$\ddot{x}(t)-\alpha\dot{x}(t)+\beta K_px(t)=0+0+\beta K_pr=\beta K_pr$$

Thus, $x_p(t)=r$ is a particular solution to the non-homogenous equation. The steady-state value of $x_p(t)$ is $r$.

\subsection*{A.3: Steady-state value of equation \ref{eq:Q4ODE}}

All steady-state values of the homogenous equation are $0$ and the steady-state solution to the particular solution of the non-homogenous equation is $r$. This means the steady-state value of equation \ref{eq:Q4ODE} is $r$.

$$\lim_{t\rightarrow\infty}x(t)=r$$

This shows that using the input function $V(t)=K_p(r-x(t))$ is effective in moving the cart from position $x=0$ to position $x=r$.

\end{document}